\documentclass[journal]{IEEEtran}
\usepackage{subfig}
\usepackage{graphicx}
\usepackage[square,numbers]{natbib}            
\usepackage{microtype}
\usepackage{amsmath} 

\ifCLASSINFOpdf
\else
\fi
\begin{document}
%
\title{Bayesian Nonparametric Models for Synchronous Brain-Computer Interfaces}
%
%
%

\author{Jaime~F~Delgado~Saa,~\IEEEmembership{Member,~IEEE,}
        Mujdat~Cetin,~\IEEEmembership{Member,~IEEE,}
\thanks{Jaime F Delgado Saa is with the Department
of Electrical and Electronics Engineering, Universidad del Norte, Colombia,
e-mail: (jadelgado@uninorte.edu.co).}
\thanks{Mujdat Cetin is with the Faculty of Natural Sciences and Engineering, Sabanci University, Turkey}}

%
%

\markboth{Research Paper}
{Bayesian Nonparametric Models for Synchronous Brain-Computer Interfaces}

%



\maketitle

\begin{abstract}
A brain-computer interface (BCI) is a system that aims for establishing a non-muscular communication path for subjects who had suffer from a neurodegenerative disease. Many BCI systems make use of the phenomena of event-related synchronization and de-synchronization of brain waves as a main feature for classification of different cognitive tasks. However, the temporal dynamics of the electroencephalographic (EEG) signals contain additional information that can be incorporated into the inference engine in order to improve the performance of the BCIs. This information about the dynamics of the signals have been exploited previously in BCIs by means of generative and discriminative methods. In particular, hidden markov models (HMMs) have been used in previous works. These methods have the disadvantage that the model parameters such as the number of hidden states and the number of Gaussian mixtures need to be fix \textit{a priori}. In this work, we propose a Bayesian nonparametric model for brain signal classification that does not require \textit{a priori} selection of the number of hidden states and the number of Gaussian mixtures of a HMM. The results show that the proposed model outperform other methods based on HMM as well as the winner algorithm of the BCI competition IV.         
\end{abstract}


%
\IEEEpeerreviewmaketitle

\section{Introduction}
A Brain Computer Interface (BCI) is a system that provides an alternative communication path for people who had suffer a disease or have had an accident that compromises their ability to perform motor tasks. Also, applications for healthy subjects in areas of multimedia and gaming started to incorporate these technologies in the recent years \cite{Nijholt2009,galway2015}. BCIs make use of the brain signals to control external devices that help the subject to communicate and interact with the environment. A significant portion of approaches to BCIs are based on the comparison of power values of the EEG signal during the execution of imaginary motor tasks. However, the well-known phenomena of Event Related Synchronization (ERS) and Event Related De-synchronization (ERD) \cite{DaSilva_1999} provide more information that can be used for improving the performance of the BCIs. This information is related not only to the difference of power of the signals during the execution of different motor tasks but also to their change in time in different frequency bands. In order to make use of this information more explicity in a statistical modeling context, algorithms such as Hidden Markov Models (HMM) \cite{Obermaier2001, Suk2010} have been used in combination with features that describe the temporal behavior of the EEG signals \cite{Hjorth1970, Vidaurre2009}. Although the use of EEG time-frequency features provide good results \cite{Magjarevic2010, Mina2006, Palaniappan2005, Mu2009, Delgado2010} in BCI applications, a combination of the time-frequency power features and algorithms that take into account the temporal dynamics of those features have been rarely reported ~\cite{Obermaier2001, Suk2010, Ozgur2010}. One possible reason for this is that selection of model parameters (states, Gaussian mixtures, etc. in HMM) and frequency bands becomes problematic. In this work, we propose the use of the signal time-frequency distribution calculated using auto-regressive models, and a Bayesian nonparametric model for classification of two different motor tasks. The proposed model selects the model order directly from data assuming initially an infinite number of hidden states and an infinite number of Gaussian mixtures for modeling of the observations. The results show that the Bayesian approach outperforms the other HHM-based approaches used for comparison and other methods that report results on the data set used in this work.

\section{HMM Approach}

A HMM is a finite automaton which contains a set of discrete states $H$ emitting a feature vector $x_t$ at each time point $t$. Given that these kind of models are generative, it is necessary to determine the joint probability over observations and labels, which requires all possible observation sequences to be enumerated. In order to make the inference problem tractable, conditional independence is assumed, meaning that future states are independent from the past states given the current. The structure of the state sequence is described by
\begin{equation}
h_t |h_{t-1} \sim \pi_{h_{t-1}}
\end{equation}

\noindent where $\pi_{h_{t-1}}$ is a state dependent transition distribution for the state $h_{t-1}$. We also have that given $h_t$, $x_t$ is independent of observations and states at different time points:

\begin{equation}
x_t |h_t \sim F(\theta_{h_t})
\end{equation}
\noindent that is, $x_t |h_t$ is distributed according to a function $F(\theta_{h_t})$ where $\theta_{h_t}$ are the emission parameters related to the state $h_t$.

The learning problem is defined as a search for the parameters that maximize the log-likelihood of the observations. These parameters are the state transition matrix $\pi$, for which each entry $\pi_{i,j}$ represents the probability to pass from state $i$ to state $j$, the vector of initial probabilities $\pi^0$ so that $ h_1 \sim \pi^0 $, and the parameters $\theta$ of the emission distribution $F$ for each state. In this work, the emission distribution for each state is modeled using Gaussian mixtures.

The joint probability density function of the observations and states is given by:
\begin{equation}
P(h_{1:n},x_{1:n}) = \pi^0(h_1)p(x_1/h_1)\prod_{t=2}^{n}{p(x_t/h_t)p(h_t/h_{t-1})}
\end{equation}

\begin{figure}[!t]
	\centering
		\includegraphics[width=3.5in,keepaspectratio = true]{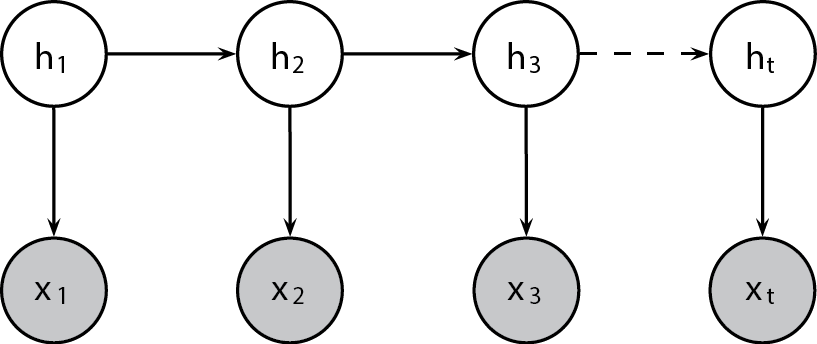}
	\caption{Graphical model representation of a HMM}
	\label{fig:HMM_GRAPH}
\end{figure}

\begin{figure}[!t]
	\centering
	\subfloat[Class one: Left hand]{\includegraphics[width=3.0in,keepaspectratio = true]{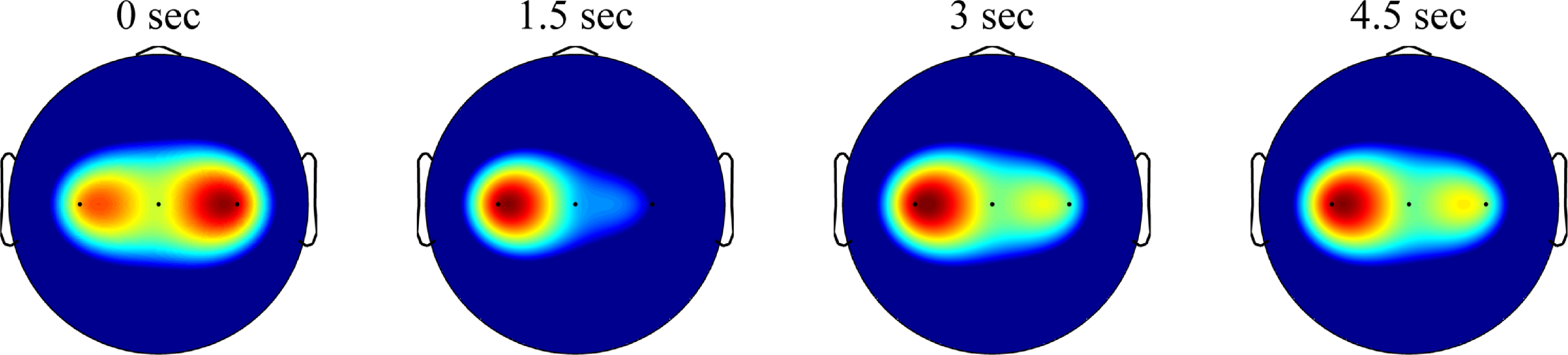}}\quad
	\subfloat[Class two: right  hand]{\includegraphics[width=3.in,keepaspectratio = true]{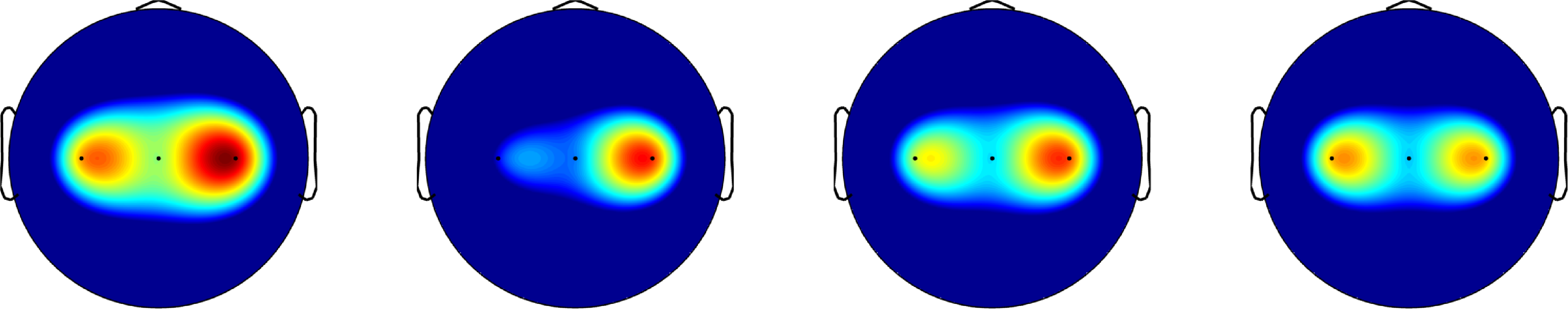}}
	\caption{Scalp topographical distribution of the power during the execution of two different imaginary motor tasks.}
	\label{fig:TP_EXAMPLE}
\end{figure}

The joint distribution of the HMM is represented by the graph in Figure \ref{fig:HMM_GRAPH}. The hidden states $h$ represent different states during the execution of specific mental tasks. Such states are related to the power levels of the signal. The assumption of these different states during the execution of the task is justified by the time course of the EEG power in specific frequency bands. Figure \ref{fig:TP_EXAMPLE} shows the average spatial distribution of the scalp EEG power in the alpha band (8-13 Hz) for one subject using bipolar montages over the electrodes locations C3, Cz and C4  at different time points, for two different classes. This reveals the dynamics (sequence of states) of the signal during the execution of different motor tasks. The sequence of these states is modeled by a HMM for each class using training data. Inference in the model is done using the \textit{forward-backward algorithm} \cite{Rabiner:1990}  which provides an efficient scheme for passing messages in the graph, making possible the calculation of marginals for problems of filtering $p(h_t|x_1,...,x_t)$, prediction $p(h_{t+\tau}|x_1,...,x_t)$ for $\tau > 0$ and smoothing $p(h_t|x_1,...,x_{\tau})$ for $\tau > t$.

In the BCI problem studied in this work, the decision about the task executed by the subject is made at the end of the trial. Therefore, the problem to be solved is one of filtering. Model parameters ($\pi$, $\theta$) are learned from training data using the Expectation-Maximization algorithm \cite{Dempster77}. During the expectation step the expected value of the likelihood of the model is calculated using the \textit{forward-backward} algorithm. During the maximization step, gradient search is used to maximize the expected value of the likelihood. 

The selection of the number of hidden states and the number of Gaussian mixtures is made in two ways: 1.) Based on the description of the task, the number of hidden states is set to three, representing start, execution, and end of the task. The number of Gaussian mixtures was fixed to two as a reasonable tradeoff between the expressive power of the model and its simplicity. 2) Three-fold cross-validation is performed using the training data, looking for a combination of the number  of hidden states and the number of Gaussian mixtures that maximizes a cost function, that in this work is selected to be classification accuracy.  

Once the number of hidden states and the number of Gaussian mixtures is fixed one model is trained for each class using all the training data available. During testing, the label assigned to each sequence is determined by calculation of the likelihood of the data on each model. The model with higher likelihood determines the assigned label.

\section{Bayesian Nonparametric HMM Approach}
The motivation for using HMM-like models is to extract temporal information from the brain signals. The underlying assumption is that during the execution of the task the rhythms produced by the brain go through a sequence of states. We link these states to changes in the power of the brain signals in specific frequency bands. However, the number of states is not known a priori. Furthermore, during each one of these states the distributions of the brain signals are complex and therefore, are better modeled by multimodal distributions. Again, the number of components involved in each multimodal distribution is unknown. The Hierarchical Dirichlet Processes HMM (HDP-HMM) with Dirichlet Processes Gaussian Mixtures \cite{Fox2010} presents a solution to such problems faced by conventional HMM approaches.

\subsection{The HDP-HMM and the Sticky HDP-HMM}
In \cite{Teh2006} a nonparametric Bayesian approach to HMM in which the HDP defines a prior distribution on transition matrices over countably infinite state spaces is presented. The HDP-HMM leads to data-driven learning algorithms which infer posterior distributions over the number of states. One serious limitation of the HDP-HMM presented by \cite{Teh2006} is that the model has a tendency to produce unrealistic rapid dynamics. In \cite{Fox2010} a modified HDP-HMM so called sticky-HDP-HMM is proposed, augmenting the original HDP-HMM by including a parameter for self-transition bias and placing a separate prior on this parameter. 

A DP denoted by DP($\gamma,H$), where $H$ is a base distribution and $\gamma$ is a concentration parameter, is a distribution over countably infinite random measures:
\begin{align}
	G_0(\theta) = \sum_{k=1}^{\infty}{\beta_k\delta(\theta-\theta_k)}&& \theta \sim H
\end{align}

\noindent on a parameter space $\Theta$. The weights $\beta_k$ are sampled via a stick-breaking construction \cite{fox2008hdp}, so that $\beta \sim$ GEM($\gamma$) where GEM($\cdot$) denotes the stick breaking construction:
\begin{align}
	\beta_k = \beta_k'\prod_{l=1}^{k-1}{(1-\beta_l')}&&\beta_k' \sim Beta(1,\gamma)
\end{align}

\begin{figure}[t!]
	\centering
		\includegraphics[width=3.5in]{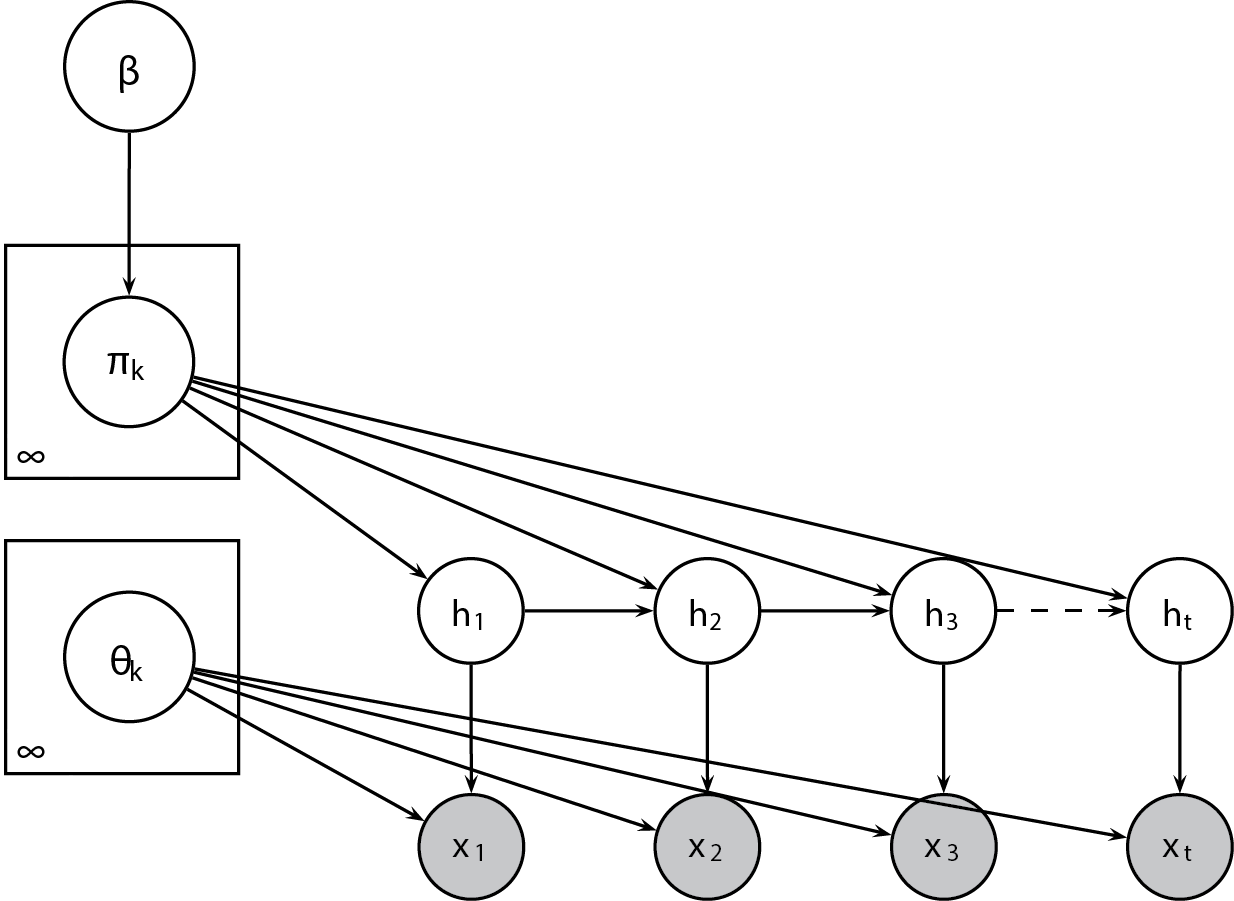}
	\caption{Sticky HDP-HMM graph}
	\label{fig:DPHMM_GRAPH}
\end{figure}

The HDP proposed by \cite{Teh2006} extends the DP to cases in which groups of data are produced by unique, generative processes \cite{Fox2010}. The HDP places a global Dirichlet process prior DP($\alpha$,$G_0$) on $\Theta$ and draws group specific distributions: 
\begin{align}
G_j(\theta)& = \sum_{k=1}^{\infty}{\pi_{jk}\delta(\theta-\theta_{jk})}&& \pi_{j} \sim DP(\alpha,\beta) 
\end{align}

\noindent where $\pi_j$ represents the transition distribution for a specific state, $h_t$ denotes the state of the Markov chain at time $t$. Figure \ref{fig:DPHMM_GRAPH} shows the graphical model for the Sticky HDP-HMM. Given the properties in the HMM the state $h_{t-1}$ indexes the group to which the observation $x_t$ is assigned and the current state $h_t$ defines the parameter $\theta_{h_t}$ used to generate the observation $x_t$. The modification proposed by \cite{Fox2010} allows to incorporate prior information that slow, smoothly varying dynamics are more likely. Unlike the original HDP-HMM, this modified version does not allow state sequences with unrealistic fast dynamics to have large posterior probability. To that end, the work in \cite{Fox2010} proposes to sample transition distributions as follows:
\begin{align}
	\pi_j \sim DP\left(\alpha+\kappa,\frac{\alpha\beta+\kappa\delta_j}{\alpha+\kappa}\right)
\end{align}   

This modification for $\kappa$ values greater than zero has the effect of increasing the prior probability of self transitions $E[\pi_{jj}]$. When $\kappa$ is equal to zero the original HDP-HMM is obtained.

\begin{figure}[t!]
	\centering
		\includegraphics[width=3.0in]{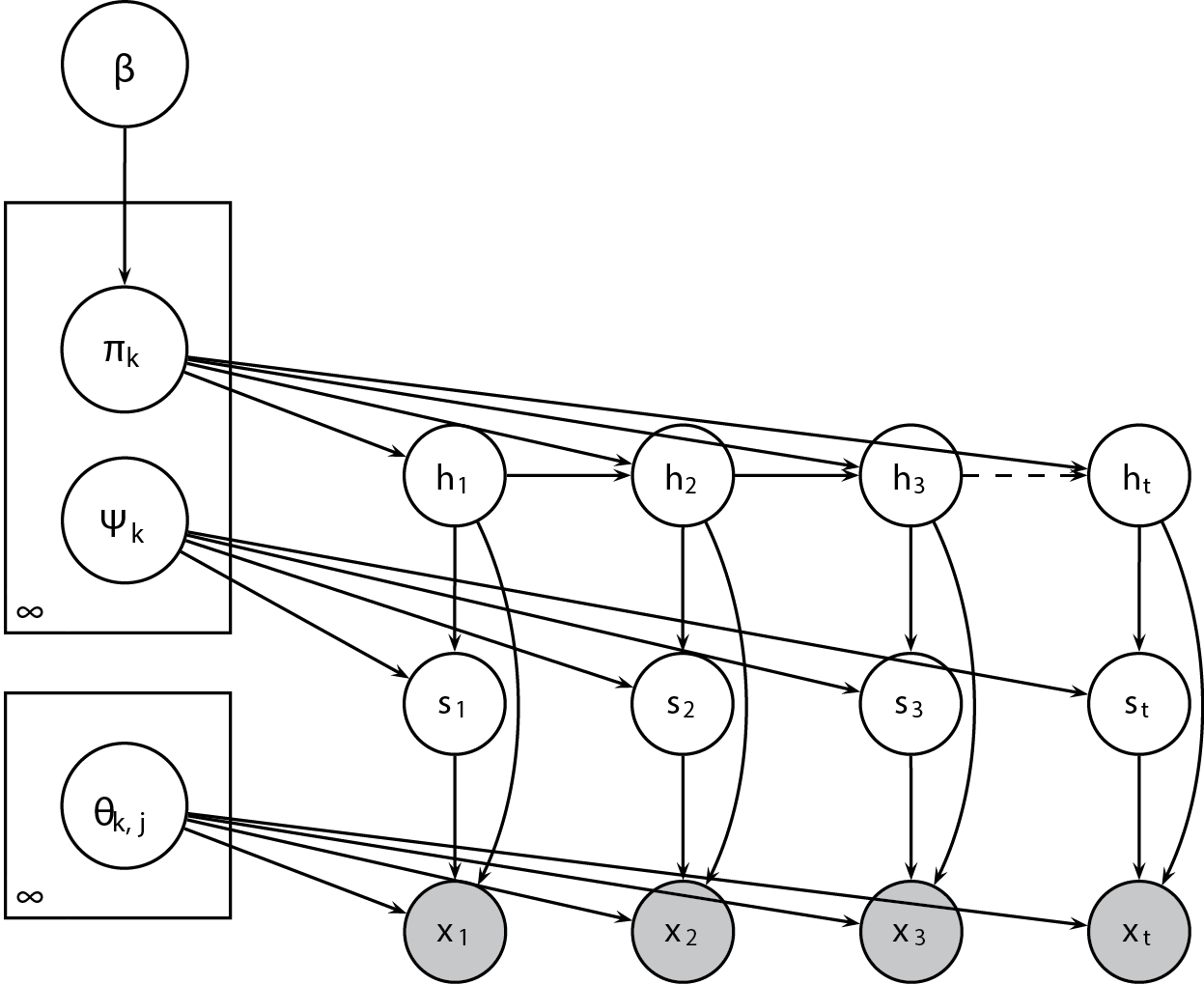}
	\caption{Sticky HDP-HMM graph with DP Gaussian mixtures }
	\label{fig:DPHMM_GRAPH_DPG}
\end{figure}

In many applications the distribution of the data associated with each hidden state is complex and is better modeled with a multimodal distribution. This motivates the use of HMM in which each hidden state is associated with a mixture of Gaussian distributions, which poses the problem of selection of the number of mixtures. Just like the model order problem associated with the number of states, this problem can be solved by using DP as well. The Sticky HDP-HMM is extended in \cite{Fox2010} by defining a DP mixture of Gaussians, including a new variable $s_{t}$ (see Figure \ref{fig:DPHMM_GRAPH_DPG}) which indexes the Gaussian mixture component of the $h_{t}-th$ hidden state. Therefore for the variables in Figure \ref{fig:DPHMM_GRAPH_DPG}, we have:

\begin{align}
	s_t \bigg| \{\psi_j\}_{j=1}^{\infty}, 	h_t \sim \psi_{h_t} \\
	x_t \bigg| \{\theta\}_{k,j=1}^{\infty},	s_t, h_{t} \sim F(\theta_{h_t,s_t})
\end{align} 

\noindent that is, $x_t$ is distribuited according to a function $F(\theta_{h_t,s_t})$, where $\theta_{h_t,s_t}$ are the emision parameters related to the state $h_t$ and mixture $s_t$.

We model the brain signals by fitting one HDP-HMM with a DP Gaussian mixtures model for each class. For the Gaussian Mixtures, a Gaussian prior is placed on the mean while an Inverse-Wishart prior is placed on the variance of the components of each mixture following \cite{Fox2010}. Classification is performed in the same fashion as for the HMM approach Section II. Using test data, calculation of the log-likelihood of the data given the model is performed. The model that is more likely to produce the data determines the declared class.

\section{Description of Experiments and Methods}

\subsection{Problem and Data Set Description}
In typical BCI applications based on the imagination of motor activity, the subject is requested to execute imaginary motor tasks following a visual cue. It is known that the imagination of motor activities produces synchronization and/or de-synchronization of the electrical signals recorded over the motor cortex and that this process has an asymmetrical spatial distribution during the imagination of the motor task (e.g., imagination of movement of a particular leg produces changes in the power of electrical signals in the contra-lateral region of the brain, see Figure \ref{fig:TP_EXAMPLE}). The models are trained using data obtained from an initial session in which the subject is requested to execute different motor tasks. Then, given some new (test) data, the task is to run an inference algorithm to perform classification of the imaginary motor task. 

\begin{figure}[t!]
\centering
\subfloat[EEG channels]{\includegraphics[width=1.0in, keepaspectratio = true]{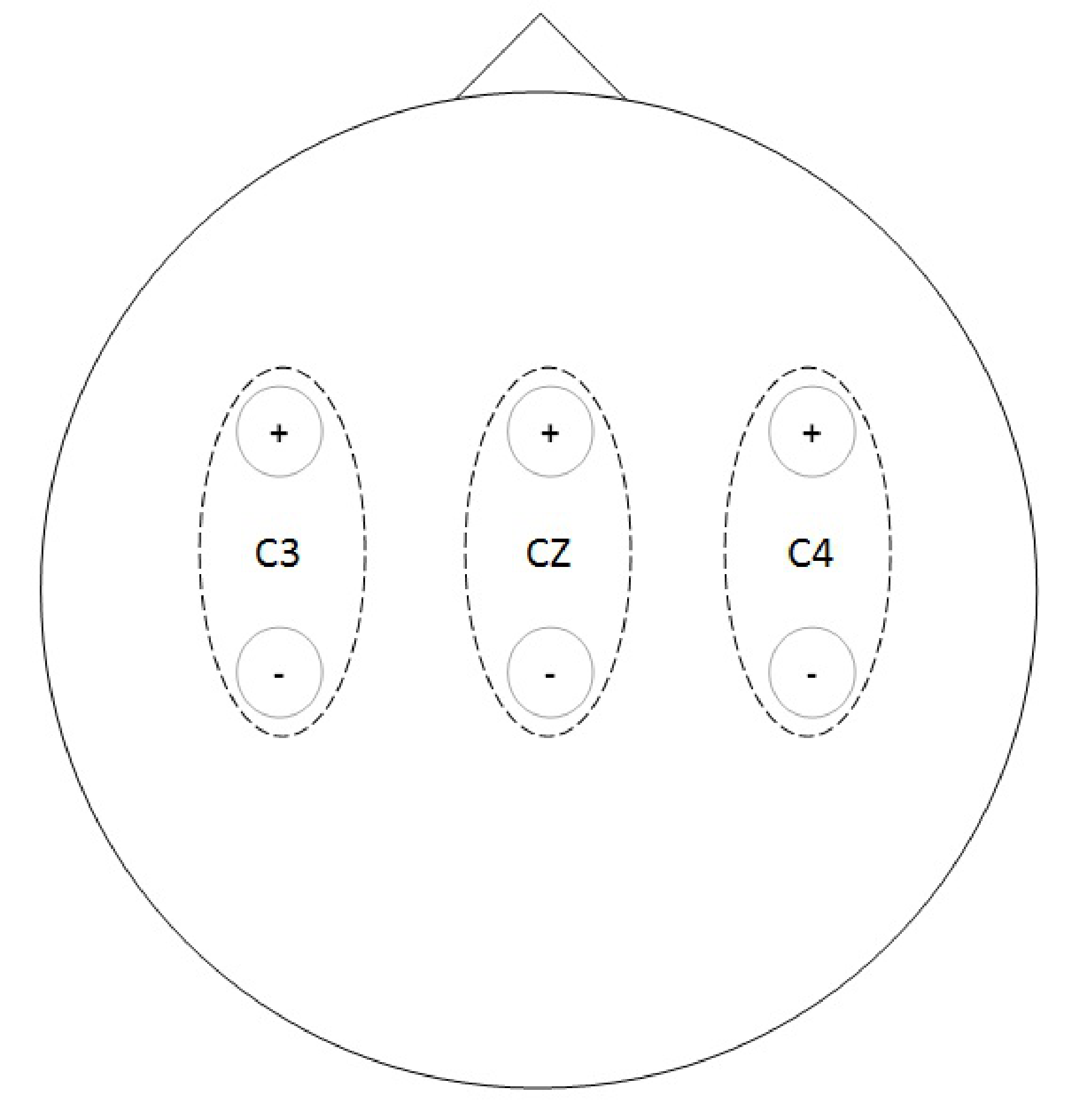}\label{fig:EEG_CHANNELS}}\qquad
\subfloat[EOG channels]{\includegraphics[width=1.0in, keepaspectratio = true]{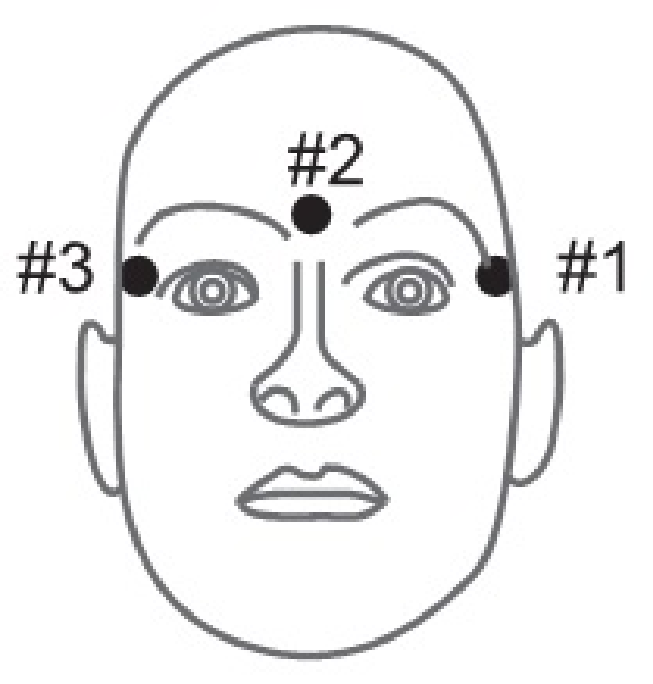}\label{fig:EOG_CHANNELS}}
\caption{Electrode positioning for the BCI competition IV data set 2b.}
\label{fig:data-set_channels}
\end{figure}

\begin{figure}[t!]
\centering
\includegraphics[width=3.5in, keepaspectratio = true]{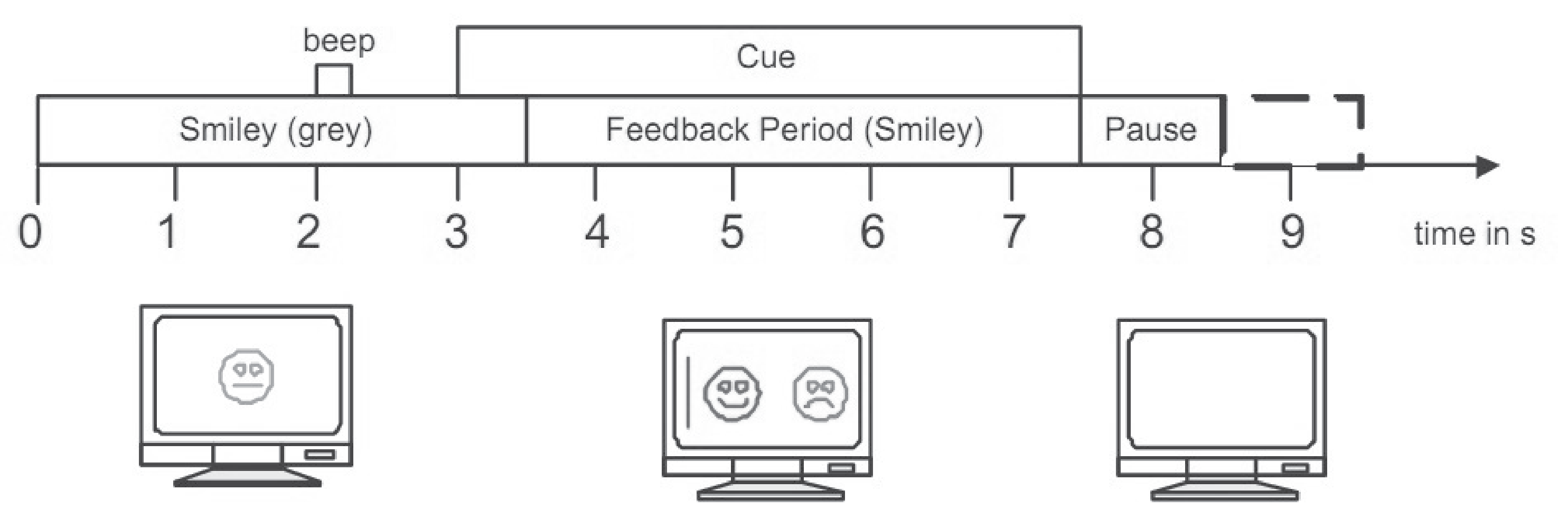}
\caption{Time scheme for the experimental procedure.}
\label{fig:scheme}
\end{figure}

In this work, Data Set 2b of BCI competition IV \cite{grazdatasetb2008} was used. The dataset contains a total of nine subjects of which two where left out because previous results show that no activity related to the task was detected for these subjects (see results of the BCI competition \cite{grazdatasetb2008}), the data contains then 7 subjects, with bipolar EEG recordings over scalp positions for electrodes C3, Cz and C4 (see Figure \ref{fig:EEG_CHANNELS}). The cue-based BCI paradigm involves two classes, represented by the imagination of the movement of the left hand and the right hand, respectively. The time scheme of the sessions is depicted in Figure \ref{fig:scheme}. At the beginning of each trial, a fixation cross and a warning tone are presented. Three seconds later, a cue (indicating left or right movement) is presented and the subject is requested to perform the imaginary movement of the corresponding hand. The data set contains five sessions, three for training and the remaining two for testing. Some of these sessions involved feedback, indicating to the subject how well the imagination of the motor task has been executed, and others do not. In our work we have used the sessions with feedback because previous work has shown that temporal behavior of the EEG signals could be modified due to the feedback influence \cite{Neuper1999a}.

The data were recorded in three different sessions in different days. The training session contains 160 trails, 80 trials for each class (imaginary right hand movement, imaginary left hand movement). The testing data consist of two sessions, each one with 160 trials equally divided between the two classes.
 
\subsection{Artifact Reduction}
Linear regression was used in order to reduce the interference of electrooculography (EOG) signals in the EEG recordings according to \cite{Schlogl2007}. EOG data recorded in three channels at electrode locations shown in Figure~\ref{fig:EOG_CHANNELS} provide a measure of the eye movements executed by the subjects. In this approach, the signal recorded by the EEG electrodes is modeled as the summation of the actual underlying EEG signal and the noise, represented by a linear combination of the EOG signals interfering into the EEG electrodes \cite{Schlogl2007}.

Before the beginning of the motor task classification sessions, subjects are requested to execute different ocular movements enabling the estimation of the artifact-free EEG signal. As a complementary step, the obtained signals are band-pass filtered in the frequency band of interest for real/imaginary motor activity (8 - 35 Hz), eliminating other sources of artifact in the low and high frequency range of the EEG recordings.

\subsection{Spatial Filter}
Spatial filters can be incorporated into the pre-processing stage by means of Common Spatial Pattern (CSP). Given a two-class classification problem related to motor task, CSP provides a method to maximize the power of the signal during the execution of the class one, while minimizing it during the execution of class two and vice versa. Given the artifact-free signal $s(n)$ obtained by the regression method described above, the spatial filtered signal $c(n)$ is obtained by:

\begin{equation}
	c(n) = s(n)W,
\end{equation}

\noindent where $W$ is a matrix for which each column represents a set of subject-specific spatial patterns. For details on the procedure for calculation of the spatial filters with CSP the reader is referred to \cite{dornhege:CSP}. The signal $s(n)$ composed of three electrodes is linearly transformed  into two CSP-filtered EEG signals $c(n)$ to be used in subsequent pre-processing stages.

\subsection{Spectral Power Estimation}
The power spectrum of the signal is computed using auto-regressive (AR) models of the signal. Burg's method for AR model estimation was used because it provides better stability than the Yule-Walker method, by minimizing the error in backward and in forward direction \cite{Stoica1997}. Once the autoregressive model is obtained we estimated the power spectrum of the signal as the frequency response of the AR model. 

For estimating the AR parameters a one-second sliding window was used over the spatial filtered signals $c(n)$. For each signal segment of one second, the model is estimated and the frequency response is obtained. The overlap of the segments was fixed to 90\% of the window length. This produces a time-frequency map for each signal. From this time-frequency representation, the features to be used in the classifiers (HMM and Sticky HDP-HMM) are selected based on physiological information of the frequency bands related to execution/imagination of motor tasks. The selected frequency bands used in this work are alpha (8-13 Hz), sigma (11-15 Hz), low beta (18-23 Hz), high beta (21-26 Hz) and low gamma (25-35 Hz). The features are calculated by taking the average power across frequency at the indicated frequency bands. The frequency resolution used was ~1 Hz.


\section{Results}

\begin{table}[t!]
	\centering
		\begin{tabular}{c c c c c}
		 \hline
			Subject 	& Chin					& HMM-FP 	&HMM-CV	& HDP-HMM\\
			\hline
			B01				& 0.40					& 0.38 			&0.43		&0.57    \\
			B04				&	0.95					& 0.93 			&0.94		&0.92		\\
			B05				&	0.86					& 0.81 			&0.86		&0.83		\\
			B06				&	0.61					& 0.61 			&0.66		&0.81    \\
			B07				&	0.56					& 0.59 			&0.63		&0.57		\\
			B08				&	0.85					& 0.79 			&0.80		&0.81		\\
			B09				&	0.74					& 0.70 			&0.71		&0.79    \\
			\hline
			Average 	&	0.71					& 0.69 			&0.72		&\bf0.76    \\
			\hline
		\end{tabular}
	\caption{Comparison of the proposed (Sticky) HDP-HMM approach with HMM and the BCI's competition winner (Chin.). HMM-FP corresponds to a HMM with parameters fixed a priori (three hidden states, Gaussian mixtures of two components per hidden state). HMM-CV corresponds to HMM with parameters selected by 3 Folds-Cross-validation. HMM-FP, HMM-CV, and Sticky HDP-HMM use the same set of features. The metric used is Cohen's Kappa .}
	\label{tab:results}
\end{table}

\begin{figure}[t!]
\centering
\includegraphics[trim= 2.5in 0 3in 0, clip, width=3.2in, keepaspectratio = true]{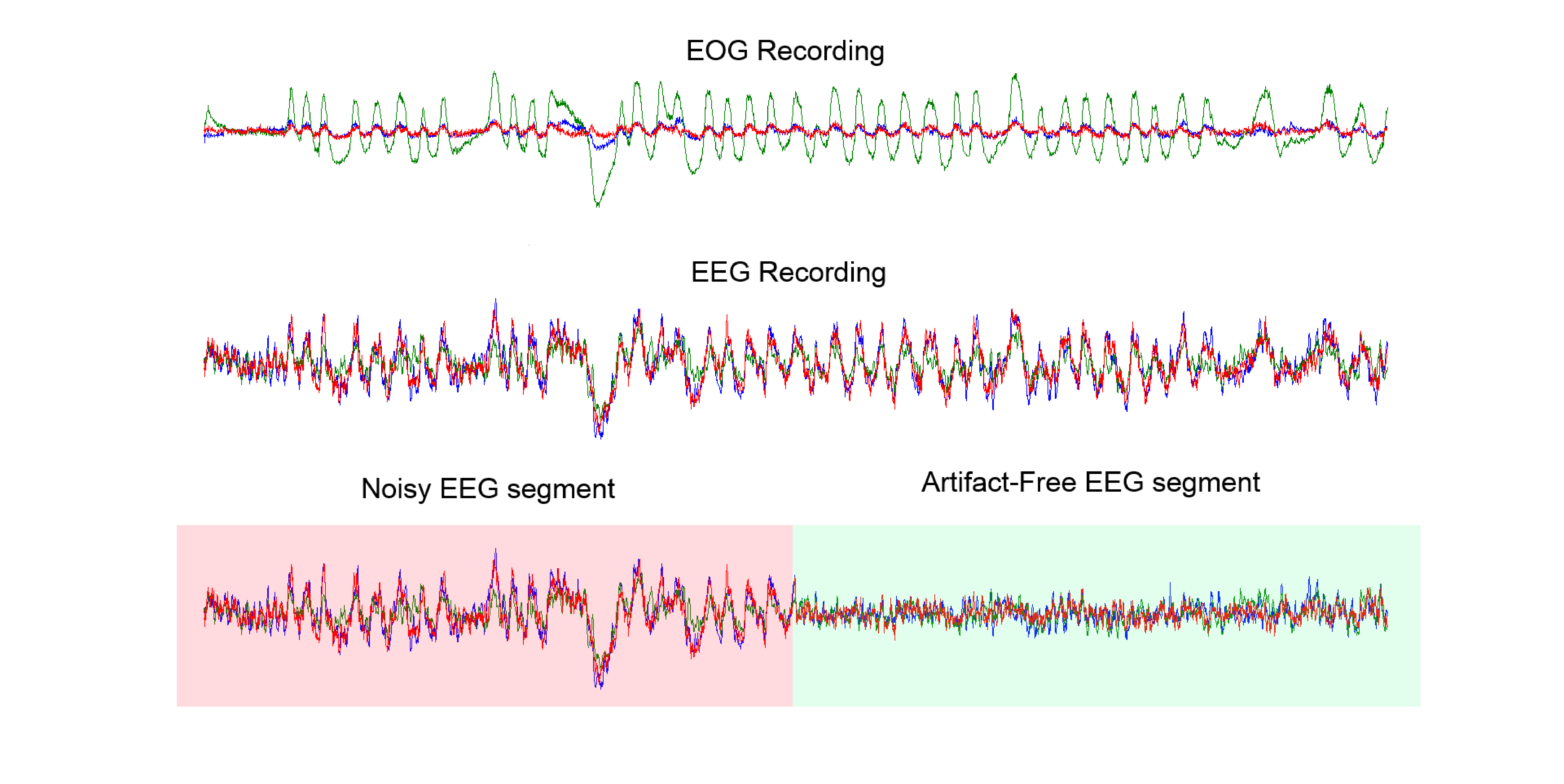}
\caption{EOG artifact removal}
\label{fig:EOG_removal}
\end{figure}

Figure \ref{fig:EOG_removal} shows an example of the artifact reduction stage. Note that the EOG interference on the EEG recordings is appreciable. However, this effect can be reduced by linear regression. The bottom plot shows a noisy segment of EEG. The red area corresponds to the portion of the EEG recordings that was used to learn the regression coefficients that explain how the EOG signals propagate through the scalp by volume conduction affecting the EEG recordings. The segment in green shows a considerable reduction of the artifacts.

\begin{figure*}[t!]
\centering
\includegraphics[width=7in, keepaspectratio = true]{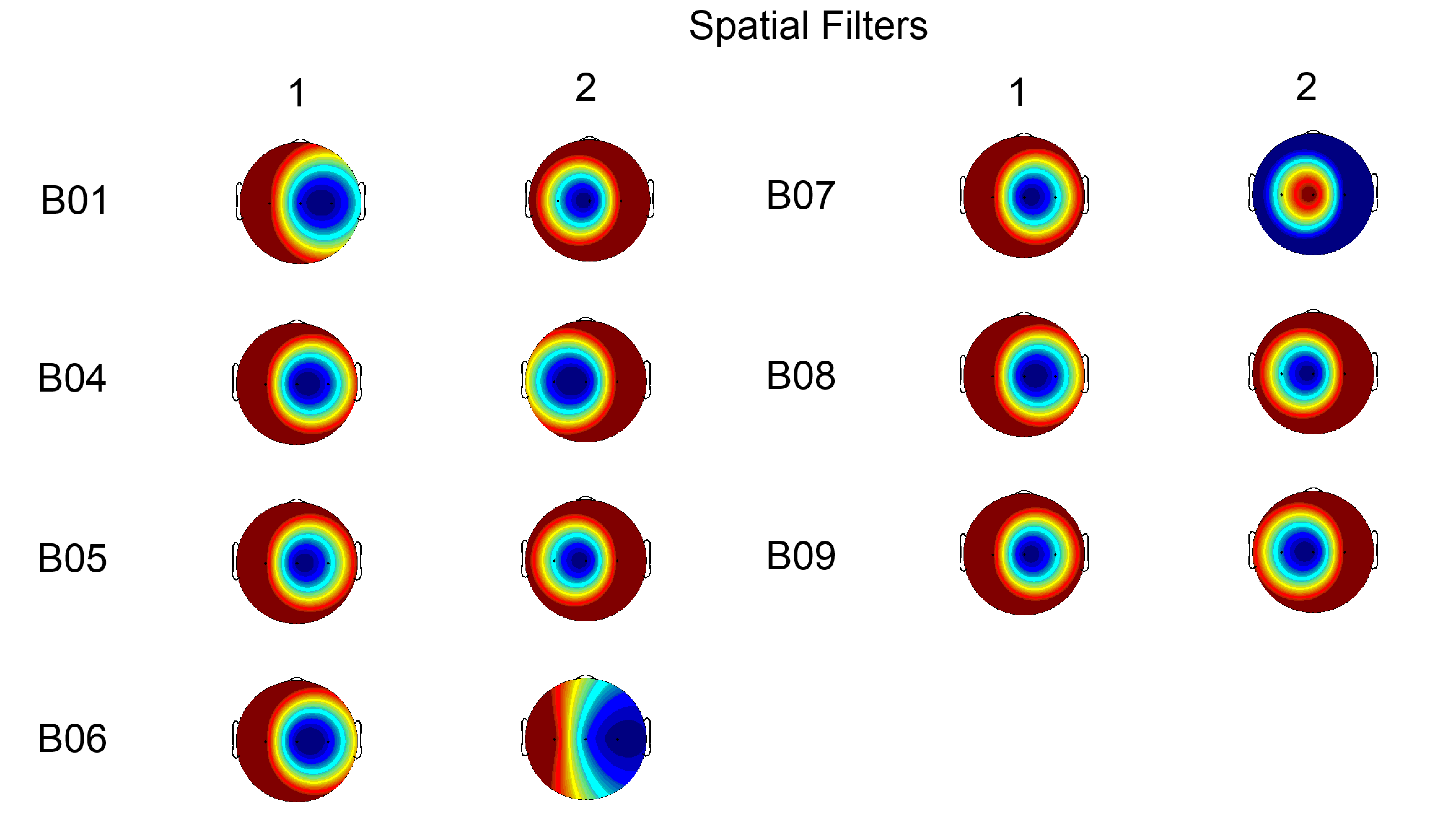}
\caption{Topographical projection of the spatial filters.}
\label{fig:CSP_TOP}
\end{figure*}

The CSPs can be projected over the scalp to visualize how the signals from different regions in the scalp contribute to form the common spatial patterns. These plots are presented in Figure \ref{fig:CSP_TOP}. Given that the positions of the electrodes in this data set represent left, right, and central areas over the motor cortex, it is easy to visualize the operation of the filters. It can be observed that for most of the subjects one of the filters gives more importance to electrode C4 (right) and the other filter gives more importance to the electrode C3 (left). This is consistent with the fact that the representation of the hand is contra-lateral, i.e., imagination of the movement of the left hand produces ERD in the right hemisphere while imagination of movement in the right hand produces ERD in the left hemisphere. According to the spatial filters in Figure \ref{fig:CSP_TOP}, the electrode Cz does not provide significant information for classification in most of the subjects.

Classification results are summarized in Table \ref{tab:results}. We use the kappa values \cite{dornhege:evaluation} as the metric for comparing different methods. The kappa values are defined according to:

 \begin{equation}
\label{eqn:kappa}
    \kappa=\frac{C\times P_{cc}-1}{C-1}
\end{equation}
\noindent where $C$ is the number of classes and $P_{cc}$ is the probability of correct classification\footnote{Equation (\ref{eqn:kappa}) takes this simple form given that the same number of samples for each class is available for each subject in each session.}. Relatively larger kappa values indicate better performance.

The average performance of the Sticky HDP-HMM is higher than the results obtained with all other methods. Also, compared to the results of the BCI competition, where the maximum kappa value reported was 0.60 the Sticky HDP-HMM provides better results. It is worth noting that the value reported by the BCI competition corresponds to the maximum value obtained along the execution of the task. In the case of the Sticky HDP-HMM and the HMM-like models presented in Table \ref{tab:results}, the time of maximum performance is expected to be at the end of the trial. The HMM with fixed parameters (HMM-FP column in Table \ref{tab:results}) makes use of three hidden states representing beginning, execution, and end of the motor task, and Gaussian mixtures of two components were allowed per each hidden state. The HMM-CV refers to the HMM approach with the number of hidden states and Gaussian mixtures learned using cross-validation. HMM-CV method uses three-fold cross-validation for selection of the number of states and the number of components of the Gaussian mixtures. The space of the parameters for the HMM-CV is 1 to 3 hidden states and 1 to 3 Gaussian mixtures.

\section{Conclusion}
In this work a Bayesian nonparametric model that makes use of hierarchical Dirichlet processes has been presented for classification of sensorimotor rhythms in synchronous BCIs. First, preprocessing methods including artifact reduction by linear regression, spatial filtering by common spatial patterns and spectral power estimation have been performed on the data. The proposed method aims to model the dynamics of the EEG signal during the execution of the motor task showing that this information can increase the performance of the system. Although methods based on HMM that aim to model the dynamics of the EEG has been introduced before, the main disadvantage of these classic HMM methods is that the number of hidden states and the parameters of the distribution that models the the data have to be set {\em a priori}. The proposed method based on HDP-HMM provides a solution for the selection of the number of hidden states and the number of components of the mixture of Gaussians used to model the distribution of the data. This is achieved by inferring a posterior distribution over the number of hidden states by means of Hierarchical Dirichlet Process, which allows an infinite number of states. Also, Dirichlet processes are used to model the number of components of the mixture of Gaussians. The results show that these methods provide good classification results. In particular, the proposed HDP-HMM approach with self bias transition (Sticky HDP-HMM) provides better results outperforming the classical HMM approaches and the method proposed by the winner of the BCI competition IV.

\section*{Acknowledgment}
This work was partially supported by Universidad del Norte (Colombia), the Scientific and Technological Research Council of Turkey under Grant 111E056, and by Sabanci University under Grant IACF-11-00889.

\ifCLASSOPTIONcaptionsoff
  \newpage
\fi

\bibliographystyle{ieeetr} 
\bibliography{Bibliography}  




\end{document}